\documentclass[sigconf]{acmart}

\AtBeginDocument{%
  \providecommand\BibTeX{{%
    Bib\TeX}}}

\copyrightyear{2026}
\acmYear{2026}
\setcopyright{acmcopyright}
\acmConference[KDD RelKD '26]{Proceedings of the International Workshop on Resource-Efficient Learning for Knowledge Discovery (RelKD 2026), co-located with KDD 2026}{Aug 10, 2026}{Jeju, Korea}
\acmBooktitle{Proceedings of the International Workshop on Resource-Efficient Learning for Knowledge Discovery (KDD RelKD '26), Aug 10, 2026, Jeju, Korea}
\acmPrice{}
\acmDOI{10.1145/nnnnnnn.nnnnnnn}

\usepackage{dblfloatfix}
\usepackage{soul}
\usepackage{url}
\usepackage{hyperref}
\hypersetup{colorlinks=true,citecolor=citeColor,urlcolor=citeColor,linkcolor=linkColor}
\usepackage{graphicx}
\usepackage{amsmath}
\usepackage{amsthm}
\usepackage{booktabs}
\usepackage{algorithmic}
\usepackage{multirow}
\usepackage{tabularx}
\usepackage{enumitem}
\usepackage{bbm}
\usepackage{xcolor}
\newtheorem{problem}{Problem}
\usepackage[ruled,linesnumbered,vlined]{algorithm2e}
\usepackage{pgfplots}
\pgfplotsset{compat=1.16}
\usepgfplotslibrary{groupplots}
\usetikzlibrary{calc}

\AtBeginDocument{%
  \providecommand\BibTeX{{%
    \normalfont B\kern-0.5em{\scshape i\kern-0.25em b}\kern-0.8em\TeX}}}

\begin{document}

\title{Brick-DICL: Dynamic In-Context Learning for Automated Brick Schema Classification}

\author{Yiyue Qian$^{1}$, Shinan Zhang$^{1}$, Huan Song$^{1}$, Negin Sokhandan$^{1}$, Hannah Marlowe$^{1}$, Diego Socolinsky$^{1}$}
\affiliation{%
  \institution{\{iamyiyue, shinanz, huanso, ngnsl, marloweh, sclinsky\}@amazon.com}
  \country{}
}
\affiliation{%
  \institution{$^{1}$Amazon AWS Generative AI Innovation Center}
  \country{}
}

\renewcommand{\shortauthors}{Qian et al.}

\begin{abstract}
Building Management Systems (BMS) are essential for optimizing energy efficiency and operational performance in modern buildings. However, the lack of standardization across BMS points from different manufacturers creates significant barriers to integration and data utilization. While the Brick schema offers a standardized ontology for building systems, mapping BMS points to appropriate Brick classes presents three critical challenges: (i) the extensive number of Brick classes (936 in the latest version), (ii) limited domain-specific knowledge in large language models (LLMs), and (iii) substantial manual effort required for verification. To address these challenges, we propose \textbf{Brick-DICL}, a two-stage dynamic in-context learning framework for automated Brick schema classification. Brick-DICL consists of two primary components: metadata-RAG, which retrieves relevant examples to enhance LLMs' domain knowledge, and class-RAG, which narrows down potential Brick classes to address the large classification space. Additionally, we implement a multi-LLM filtering mechanism that compares predictions across multiple models, flagging low-confidence classifications for human review. As a result: (i) \textit{General}: Brick-DICL is applicable to any building management system regardless of manufacturer or metadata format; (ii) \textit{Novel and Powerful}: as the first dynamic in-context learning approach for Brick schema classification, Brick-DICL achieves significant classification accuracy improvements on building datasets, outperforming existing methods; (iii) \textit{Efficient}: our multi-LLM filtering strategy reduces manual verification effort, enabling rapid digital building onboarding. Extensive experiments demonstrate Brick-DICL's effectiveness across diverse building datasets, accelerating the path toward standardized, interoperable building management systems.
\end{abstract}


\maketitle

\section{Introduction}
\begin{figure*}[t]
 	\centering
 \includegraphics[width=0.9\linewidth]{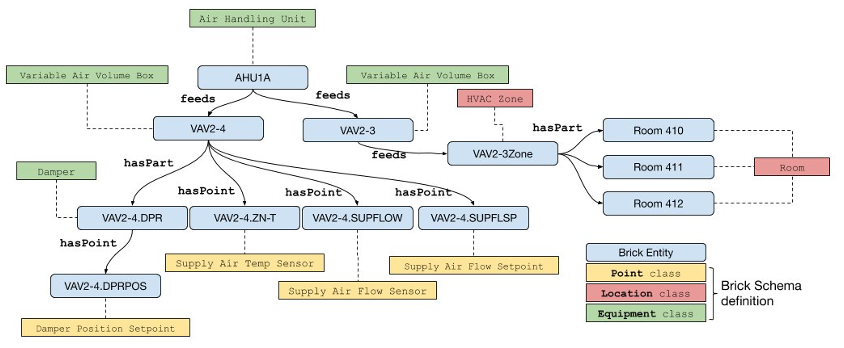}
		\vspace{-0.05in}
	\caption{An example of Brick schema hierarchy, illustrating relationships among equipment, locations, and point classes in BMS.}
	\label{fig: sample}
\end{figure*}

Building Management Systems (BMS) are vital for optimizing energy efficiency, operational performance, and sustainability in modern buildings~\cite{katib2016integrating,lee2017data,ding2019metadata,volk2014building}. However, the lack of standardization across metadata from different manufacturers creates significant barriers to seamless integration and onboarding of buildings onto digital platforms. This fragmentation complicates the collection, consolidation, and extraction of reliable data for insights and analytics, hindering efforts to achieve efficient building management~\cite{balaji2016brick}. To address these challenges, the Brick schema~\cite{balaji2016brick}, a standardized open-source ontology, has emerged as a promising solution. It provides a structured framework for representing relationships between building assets, systems, and devices, as illustrated in Figure~\ref{fig: sample}, enabling faster delivery of business outcomes through streamlined data access. The Brick schema's hierarchical graph structure, where classes are organized in parent-child relationships forming a directed acyclic graph, naturally lends itself to graph-based representation techniques for capturing class similarities~\cite{vittori2025bim}. For example, Figure~\ref{fig: sample} shows how an Air Handling Unit (AHU1A) is connected to Variable Air Volume Boxes (VAV2-4, VAV2-3), which in turn are linked to rooms and associated sensors and setpoints, illustrating the hierarchical and relational structure of the Brick schema.

Despite its advantages, mapping BMS points to Brick schema classes remains a daunting task due to two primary challenges: the vast number of Brick classes (936 classes in the least version) and the limited domain-specific knowledge of large language models (LLMs). The sheer scale of point classification demands sophisticated techniques to ensure accurate mapping across diverse building systems~\cite{steiniger2008approach,wang2018automated}. Meanwhile, LLMs, despite their impressive general capabilities, often struggle with domain-specific tasks like BMS point classification due to insufficient exposure to specialized datasets and contextual knowledge~\citep{hager2024evaluation,ge2023openagi,xue2024domain,welz2024enhancing,qian2025enhancing}. Moreover, manually verifying the point mapping classification results generated by LLMs is labor-intensive and time-consuming, particularly when dealing with thousands of points across multiple buildings~\cite{wagner2024towards,ma2024specgen,li2025verification,qian2026collabeval}.

To handle the above challenges, this paper introduces \textbf{Brick-DICL}, for \underline{\textbf{Brick}} schema classification, a novel two-stage \underline{\textbf{D}}ynamic \underline{\textbf{I}}n-\underline{\textbf{C}}ontext \underline{\textbf{L}}earning (DICL) approach that leverages retrieval augmented generation (RAG) techniques. Our approach addresses each challenge systematically: First, to manage the vast number of Brick classes, we implement a class-specific RAG that retrieves only the most relevant classes, dramatically reducing the classification space. Second, to overcome limited domain knowledge in LLMs, we develop a metadata RAG that retrieves similar, correctly-classified examples from a curated database, providing the model with context-specific knowledge for each input point. For each classification task, we first retrieve relevant examples from the metadata RAG to guide the LLM's initial prediction, then use this prediction to retrieve related Brick classes, significantly narrowing the potential classification options. Finally, to reduce manual verification efforts, we implement a multi-LLM framework with a sophisticated filtering mechanism that identifies discrepancies between model predictions, allowing human experts to focus exclusively on ambiguous or low-confidence classifications. Our framework was evaluated on two building datasets, demonstrating significant improvements in both classification accuracy and efficiency over baseline models. To conclude, our approach incorporates three key innovations:

\begin{itemize}[leftmargin=0.1in]\setlength{\itemsep}{0.2pt}
\item \textbf{General}: We introduce Brick-DICL, a general dynamic in-context learning framework, encompassing metadata RAG and class RAG, to map points to Brick classes. This general framework is applicable to points in any BMS.  

\item \textbf{Novel and Powerful}: To the best of our knowledge, Brick-DICL is the first approach that designs dynamic in-context learning on LLMs to handle the large number of Brick classes and lacking of domain knowledge issues. Brick-DICL achieves significant improvements in mapping points in BMS to Brick classes over two building datasets compared to existing methods.  

\item \textbf{Efficient}: Brick-DICL is designed to minimize human effort in verification by using multi-LLMs to validate classifications and automatically flagging low-confidence cases for human review.
\end{itemize}

\section{Related Work}
\noindent\textbf{Brick Point Classification.} BMS points serve as the input/output elements for buildings, allowing software to read sensor values and write control commands~\cite{bhattacharya2015short,hong2013towards,gao2015data,fiorelli2023automated}. However, BMS point labels are largely ad-hoc constructions driven by vendor- or site-specific conventions, creating challenges for standardization and interoperability. Balaji et al. ~\cite{balaji2018brick} introduced the Brick schema as a comprehensive solution for representing building components and their relationships, providing a foundation for standardized point classification while acknowledging the complexity introduced by the extensive class hierarchy. 
The Brick schema~\cite{balaji2018brick,bhattacharya2015short,hong2013towards,gao2015data} addresses these challenges by providing a standardized ontology for building metadata. It transforms semi-structured information implied in point labels into explicit, standardized models accessible to software.
As noted by Balaji et al., the latest Brick schema contains 936 distinct classes, creating a large-scale multi-class classification problem that is difficult to address with traditional machine learning approaches. 
The large volume of classes in the Brick schema creates a fundamental challenge for classification systems, particularly when dealing with limited training data and the need for high accuracy across all classes. Given that the Brick schema is inherently a hierarchical graph with imbalanced class distributions, some works have explored graph-based approaches to model building ontology structures and inter-class relationships~\cite{vittori2025bim,qian2022co,qian2024dual,qian2023universal,qian2025adaptive,ma2025adaptive,qian2024graph,ye2020alpha,ye2020community,zhang2020dstyle,ma2026bhygnn+,qian2022rep2vec,wen2024gcvr,wen2022disentangled,qian2021distilling,qian2021adapting,zhang2022adapting,ma2023hypergraph,ma202586hypergraph,qian2022malicious,ma2026opbench}. Inspired by the graph-based hierachy structure of the Brick schema, in this paper, we propose dynamic ICL, as it can adaptively retrieve relevant examples and narrow down the potential class space, making the classification task more manageable while maintaining high accuracy. To this end, we propose dynamic ICL, as it can adaptively retrieve relevant examples and narrow down the potential class space, making the classification task more manageable while maintaining high accuracy. 

\noindent\textbf{In-context Learning in LLMs.}
In-context learning (ICL), also known as few-shot learning in LLMs, has emerged as a transformative capability in LLMs, enabling them to adapt to new tasks without parameter updates~\cite{brown2020language,coda2023meta,weiemergent,peng2023instruction,zhao2021calibrate,agarwal2024many,coda2023meta,song2025learning}. Unlike traditional fine-tuning methods that modify internal weights, ICL allows models to learn from examples embedded directly within prompts, offering significant advantages in flexibility, computational efficiency, and effectiveness with limited examples.
Specifically, Brown et al.~\cite{brown2020language} first demonstrated this capability in their seminal GPT-3 paper, showing that LLMs can perform new tasks by conditioning on a few examples within the prompt, without any gradient updates. 
This few-shot learning paradigm represented a fundamental shift in how models could be adapted to downstream tasks. 
Despite its advantages, ICL faces several challenges: it suffers from scalability issues when dealing with complex tasks, example sensitivity where results depend on the quality of examples provided, and inconsistencies in responses. Peng et al. ~\cite{peng2023instruction} explored the limitations of ICL in specification-heavy tasks, finding that ICL performance falls short in tasks requiring extensive specifications. Recent works have explored LLM-empowered approaches for domain-specific classification~\cite{ma2025llm}, multi-agent systems for data collection~\cite{ma2025autodata}, and inference-time model steering techniques~\cite{xue2026inference,ma2026nonmonotonic}. These limitations highlight the need for more sophisticated approaches to ICL. To this end, we propose dynamic ICL in this paper to adaptively select relevant examples.

\begin{figure*}[t]
 	\centering
 	\includegraphics[width=1.0\textwidth]{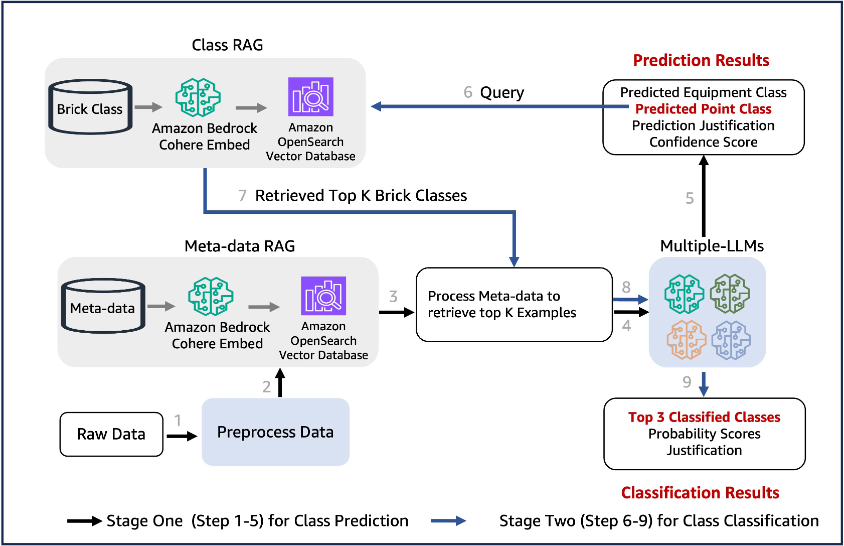}
         	\caption{The framework of Brick-DICL: (a) Raw data is first preprocessed to extract and standardize meta-data and candidate brick classes. (b) In Stage One (steps 1–5), meta-data is embedded and indexed using an embedding model and a vector database, enabling retrieval of the top-K most relevant examples for each query. (c) In Stage Two (steps 6–9), a class-level retrieval-augmented generation (RAG) process embeds and indexes candidate brick classes, retrieves the top-K classes, and combines them with the meta-data context. (d) Multiple LLMs then perform the top three classified classes with their probability scores and explanations. 
        } 
 	\label{structure}
 \end{figure*}

\section{Preliminary}
\noindent\textit{Definition 3.1.} \noindent\textbf{BMS Points and Brick Schema.}
A Building Management System (BMS) point $p$ is a fundamental data entity in building automation systems, representing a specific operational data point. Each BMS point is described by a set of attributes $A_p = \{a_1, a_2, \ldots, a_n\}$, such as equipment information, point information, and other metadata. However, these attributes are often provided in non-standardized formats that vary across vendors and systems.

To address this heterogeneity, the Brick schema provides a standardized ontology for representing building components, systems, and their relationships. It defines a set of classes $\mathcal{C} = \{c_1, c_2, \ldots, c_K\}$, where $K = |\mathcal{C}|$ is the total number of classes (936 in the latest version). These classes are organized in a hierarchical structure, with each class $c_i$ potentially having one or more parent classes, forming a directed acyclic graph. By mapping BMS point attributes to standardized Brick classes, the schema enables consistent representation of building metadata.


\noindent\textit{Definition 3.2.} \textbf{In-Context Learning (ICL).} ICL is a technique where a language model learns to perform a task by conditioning on examples provided within the prompt, without updating its parameters. Given a set of input-output examples $\{(x_1, y_1), (x_2, y_2), \ldots, (x_m, y_m)\}$ and a new input $x_{new}$, the model predicts the output $y_{new}$ by inferring patterns from the provided examples. The performance of ICL depends on the quality and relevance of the examples.

\noindent\begin{problem}
\textbf{Brick Point Classification}. Given a set of BMS points $P = \{p_1, p_2, \ldots, p_N\}$, where each point $p_i$ is described by a set of metadata attributes $A_{p_i}$, and a set of Brick schema classes $\mathcal{C} = \{c_1, c_2, \ldots, c_K\}$, the goal is to develop a classification system that maps each BMS point $p_i$ to its corresponding Brick class $c_j \in \mathcal{C}$. Formally, the task is to learn a mapping function
$f: P \rightarrow \mathcal{C}, \quad \text{where} \quad f(p_i) = \underset{c_j \in \mathcal{C}}{\arg\max} \; P(c_j \mid A_{p_i})$
such that for each BMS point $p_i$, the most appropriate Brick class $c_j$ is assigned based on its metadata attributes.
\end{problem}

\section{Proposed Model}
In this section, we present the details of Brick-DICL.
The framework consists of two main components: two-stage dynamic ICL and multi-LLM mechanism for filtering low-confidence generations, as illustrated in Figure~\ref{structure}.

\subsection{Two-Stage Dynamic In-Context Learning}
First of all, we preprocess various metadata attributes associated with BMS points. After preprocessing BMS metadata, we feed metadata to LLMs to map points in BMS into Brick classes. As discussed, traditional ICL enables LLMs to perform tasks by conditioning on examples provided within the prompt without parameter updates. However, standard ICL faces limitations when applied to domain-specific tasks with large label spaces, such as Brick point classification with 936 distinct classes. The fixed set of examples used in traditional ICL may not provide sufficient context for accurate classification across the diverse range of BMS points. To this end, we introduce dynamic in-context learning (DICL) to provide relevant examples dynamically. Specifically, 
given a set of points $\mathcal{D} = \{(x_1, y_1), (x_2, y_2), \ldots, (x_m, y_m)\}$, a retrieval function $R(x_\text{new}, \mathcal{D}, k)$ that selects $k$ examples most similar to a new input $x_\text{new}$, and a LLM model $\mathcal{M}$, DICL is formulated as:

\begin{equation}
\hat{y}_\text{new} = \mathcal{M}(\text{Instruction} \oplus \{(x_i, y_i) | (x_i, y_i) \in R(x_\text{new}, \mathcal{D}, k)\} \oplus x_\text{new}),
\end{equation}

where $\oplus$ denotes text concatenation. Here $\hat{y}_\text{new}$ is the classification that language model generated based on the dynamically constructed prompt. 
\subsubsection{Stage One: Metadata Dynamic ICL}
The first stage of our approach focuses on producing point Brick classes using metadata-based RAG (Metadata RAG). This stage lays the foundation for the more fine-grained classification in stage two.

\textbf{Metadata Retrieval.}
We first create a vector database of metadata for our training examples, enabling efficient similarity-based retrieval. Specifically, for each BMS point $p$ with attributes $A_p$, we compute an embedding vector $e_p$ using an embedding model, where $e_p = \mathcal{M}(A_p)$.
With embedding vector $e_p$, given a new BMS point $p_\text{new}$ with attributes $A_{p_\text{new}}$, we retrieve the $k$ most similar examples using cosine similarity between embedding vectors:

\begin{equation}
R(p_\text{new}, \mathcal{D}, k) = \text{argmax}_{p_i \in \mathcal{D}}^k \frac{e_{p_\text{new}} \cdot e_{p_i}}{||e_{p_\text{new}}|| \cdot ||e_{p_i}||},
\label{eq:prompt}
\end{equation}

where $\text{argmax}^k$ returns the $k$ examples with the highest similarity scores and $\mathcal{D}$ is the point set.

\textbf{Prediction Task.}
With the retrieved example $r \in R(p_{\text{new}}, \mathcal{D}, k)$, the prompt $P_{\text{stage1}}$ at the stage one in metadata DICL can be formulated as :

\begin{equation}
P_{\text{stage1}} = \text{Instruction} \oplus \left(\bigoplus_{i=1}^k \phi(r_i)\right) \oplus \psi(A_p),
\label{eq:example}
\end{equation}
where $\phi(r_i)$ formats the retrieved example $r_i$, and $\psi(A_p)$ formats the input attributes of the new BMS point. The operator $\oplus$ combines all components into a single input prompt for the language model. The retrieved examples $r$, obtained from metadata RAG in Equation~\ref{eq:example}, provide domain-specific context by demonstrating how similar attribute combinations map to specific Brick classes. These examples are dynamically selected based on cosine similarity between embedding vectors, ensuring relevance to the input attributes $A_p$.
Given the constructed prompt $P_{\text{stage1}}$, the language model $\mathcal{M}$ generates predictions as
$(c_{\text{brick}}^{(1)}, j, s) = \mathcal{M}(P)$,
where  $c_{\text{brick}}^{(1)}$ is the predicted point class at stage one, $j$ is the justification for the predictions, and $s$ is the confidence score. The justification $j$ provides interpretability by explaining how the model arrived at its predictions, while the confidence score $s$ quantifies the model's certainty. This formulation ensures that predictions are contextually grounded in relevant examples while maintaining consistency with Brick schema constraints. The structured outputs $(c_{\text{brick}}, j, s)$ are then passed to stage two for further refinement via class-specific retrieval-augmented generation.

\subsection{Stage Two: Class Dynamic ICL}  
The second stage refines the initial predictions through class-specific RAG (Class RAG). This stage addresses the challenge posed by the extensive Brick schema (936 classes) by dynamically focusing on a semantically relevant subset of classes, guided by stage one's predictions.

\subsubsection{Class-Specific Retrieval}  
Given the predicted class $c_{\text{brick}}^{(1)}$ from stage one, we retrieve a focused subset of Brick classes $\mathcal{C}_{\text{ret}} \subset \mathcal{C}$ using the class similarity. For each class $c_j \in \mathcal{C}$,  the retrieval score for class $c_j$ is calculated as 
$\text{sim}(c_j, c_{\text{brick}}^{(1)}) = \frac{e_{c_j} \cdot e_{c_{\text{brick}}^{(1)}}}{\|e_{c_j}\| \|e_{c_{\text{brick}}^{(1)}}\|}$, where $e_{c_j}$ is the class embedding using the same embedding model. 
Then the top-$m$ classes are selected as $\mathcal{C}_{\text{ret}} = \underset{c_j \in \mathcal{C}}{\text{argmax}}^m \ \text{sim}(c_j, c_{\text{brick}}^{(1)})$,
where $m$ (i.e., $m= 20$) is the hyper-parameter to control the number of retrieved classes, which can ensure reduction in classification complexity while maintaining high performance.

\subsubsection{Two-Step LLM Generation}  
Once the top $m$ (i.e., 20) classes are retrieved, we further refine them through a two-step evaluation process designed to progressively reduce difficulty.

Specifically, at the first step, the LLM evaluates all 20 retrieved classes and selects 5 preliminary candidates based on their alignment with input attributes and the retrieved examples. This step reduces complexity by narrowing down to a smaller set of plausible candidates while retaining flexibility to capture ambiguous or less obvious matches. 

At the second step, the LLM performs a more focused evaluation of these 5 candidates to select the top 3 most likely Brick classes. This step further simplifies decision-making by concentrating on a manageable number of options, allowing for detailed reasoning about fine-grained distinctions. For example, among Discharge\_Air\_Temperature\_Sensor, Supply\_Air\_Temperature\_Sensor, and Return\_Air\_Temperature\_Sensor, the LLM can prioritize Discharge\_Air\_Temperature\_Sensor based on its alignment with equipment type (e.g., RTU discharge air system).

The progressive reduction ensures that each phase is easier for the LLM to handle, as it focuses on increasingly smaller subsets of candidates. This design reduces cognitive load for the model, leading to more accurate predictions. 
The stage two can be formulated by integrating retrieved classes, retrieved examples from stage one, and sample input information:
\begin{align}
\left(\{(c_1, p_1), (c_2, p_2), (c_3, p_3)\}, j, s\right) 
  &= \mathcal{M}\big( \text{Instruction} \oplus \phi(\mathcal{C}_{\text{ret}}) \notag \\
  &\phantom{=} \oplus \left( \bigoplus_{i=1}^k \phi(r_i) \right) \oplus \psi(A_p) \big).
\end{align}
The output includes probabilities for each of the top three classes. Probabilities $p_i$ are normalized such that $\sum p_i = 1$ . The justification  $j$ explains how input attributes align with each predicted class and $s$ is the confidence score for the classification generated by the LLM.

\subsection{Multi-LLM Mechanism for Low-Confidence Generation Filter}  
To further enhance the reliability of the proposed approach and minimize human intervention, we incorporate a multi-LLM mechanism to filter low-confidence predictions. This mechanism uses multiple LLMs to generate classifications across both stages and applies a filtering strategy to identify and flag low-confidence generations. By comparing predictions from multiple LLMs, this approach ensures robustness and reduces errors caused by individual model biases or uncertainties.
Specifically, for a BMS point $p$, let $c_{\text{brick}, M_i}^{(1)}$ represent the stage one prediction from model $M_i$, and $\{c_{1,M_i}, c_{2,M_i}, c_{3,M_i}\}$ represent the top three predictions from stage two for model $M_i$. Four distinct filtering strategies are employed to assess confidence levels:
\begin{itemize}[leftmargin=0.1in]\setlength{\itemsep}{0.2pt}
\item \textbf{All Agreement}: This strategy flags a generation as low-confidence if any of the predicted results from stage one or classified results from stage two differ across the LLMs:
   \begin{equation}
   c_{\text{brick}, M_i}^{(1)} \neq c_{\text{brick}, M_j}^{(1)} \ \lor \ \{c_{1,M_i}, c_{2,M_i}, c_{3,M_i}\} \neq \{c_{1,M_j}, c_{2,M_j}, c_{3,M_j}\}.
   \end{equation}
   If all predictions across both stages are identical, the generation is considered high-confidence.

\item \textbf{Top-3 Consensus}: This strategy flags a generation as low-confidence if there is no complete overlap among the top three predicted classes across all LLMs:
   \begin{equation}
   \bigcap_{M_i \in \mathcal{M}} \{c_{1,M_i}, c_{2,M_i}, c_{3,M_i}\} = \emptyset.
   \end{equation}

\item \textbf{Top-1 Agreement}: This strategy flags a generation as low-confidence if the top one predicted class differs across any of the LLMs:
   \begin{equation}
   \exists M_i, M_j \in \mathcal{M} \ | \ c_{1,M_i} \neq c_{1,M_j}.
   \end{equation}

\item \textbf{Any-2 Consensus}: This strategy flags a generation as low-confidence if fewer than two common predicted classes exist among the top three predictions from multiple LLMs:
   \begin{equation}
   \max_{c \in \mathcal{C}} \sum_{M_i \in \mathcal{M}} \mathbb{I}(c \in \{c_{1,M_i}, c_{2,M_i}, c_{3,M_i}\}) < 2.
   \end{equation}
\end{itemize}
These strategies provide varying levels of strictness in assessing confidence. The All Agreemen strategy is the most stringent, requiring complete consistency across all models and stages, while Any-2 Consensus is more tolerant, allowing for some disagreement among models. 

Our filtering mechanism evaluates these predictions using one of the four strategies described above. Flagged generations are routed to human reviewers through an interface that aggregates all model outputs and highlights discrepancies for validation.
The multi-LLM mechanism significantly enhances classification reliability by leveraging consensus among models to identify uncertain or inconsistent predictions. By offering multiple filtering strategies with varying levels of strictness, it allows flexibility based on application requirements. 
For clarity, the pseudo-code is outlined in Alogrithm ~\ref{alg: code}.

This section shows the pseudo-code of Brick-DICL, which includes two main modules: two stage DICL, and multi-LLM for low-confidence generation filter.~\label{code}
\begin{algorithm}[h]
\renewcommand{\algorithmicrequire}{\textbf{Require}}
\renewcommand{\algorithmicensure}{\textbf{Output}}
\caption{Brick-DICL Classification Pipeline}\label{alg:code}
\begin{algorithmic}[1]
    \REQUIRE{BMS point data $\mathcal{D}$, Brick classes $\mathcal{C}$, LLMs $\mathcal{M}$, Training examples $\mathcal{T}$}
    \ENSURE{Classified BMS points with Brick classes $\{(p_i, c_{\text{final}})\}$}
    
    \STATE \textbf{Data Standardization:}
    \FOR{each $p \in \mathcal{D}$}
        \STATE Normalize metadata attributes
        \STATE Generate missing descriptions using $\mathcal{M}_{\text{desc}}$
    \ENDFOR
    
    \STATE \textbf{Build RAG Databases:}
    \STATE Meta-data RAG: Create embedding vectors for training examples
    \STATE Class RAG: Create embedding vectors for Brick class definitions
    
    \textbf{\textit{Classification Process}:}
    \FOR{each BMS point $p_i \in \mathcal{D}$}
        \STATE \textbf{Stage One - Metadata ICL:}
        \STATE Retrieve top-$k$ similar examples from Meta-data RAG
        \STATE Construct dynamic prompt with examples and input attributes
        \STATE Generate initial Brick class prediction $c_{\text{brick}}^{(1)}$
        
        \STATE \textbf{Stage Two - Class ICL:}
        \STATE Retrieve top-$m$ relevant classes from Class RAG
        \STATE Construct refinement prompt with class definitions
        \STATE Generate top-3 predictions $\{(c_1,p_1),(c_2,p_2),(c_3,p_3)\}$
        
        \STATE \textbf{Multi-LLM Filtering:}
        \FOR{each LLM $M_j \in \mathcal{M}$}
            \STATE Get predictions $(c_{\text{brick},j}^{(1)}, \{c_{1,j},c_{2,j},c_{3,j}\})$
        \ENDFOR
        \STATE Apply filtering strategies (All/Top3/Top1/Any2)
        \IF{low-confidence flagged}
            \STATE Route to human validation interface
        \ENDIF
    \ENDFOR
    
    \textbf{\textit{Human Validation}:}
    \FOR{each flagged prediction}
        \STATE Present multi-LLM predictions and justifications
        \STATE Store expert-validated results
    \ENDFOR
    
    \STATE \textbf{Return} Final classifications with confidence scores
\end{algorithmic}~\label{alg: code}
\end{algorithm}

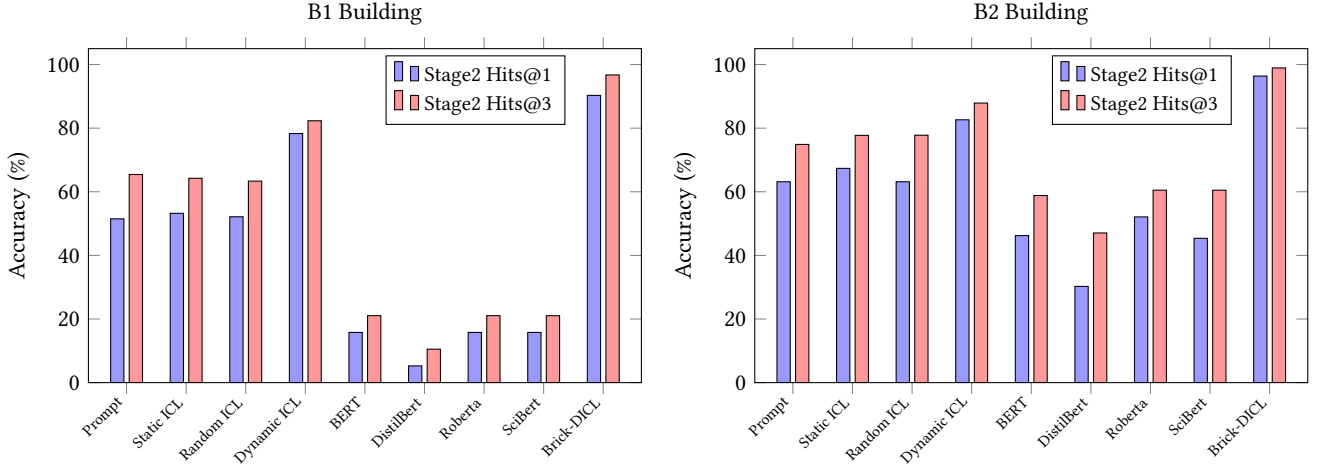
\begin{figure*}[t]
\centering
\begin{tikzpicture}
\begin{groupplot}[
    group style={
        group size=2 by 1,
        horizontal sep=1.5cm,
    },
    width=0.5\textwidth,
    height=6cm,
    ylabel={Accuracy (\%)},
    ymin=0, ymax=105,
    ytick={0,20,40,60,80,100},
    symbolic x coords={Prompt, Static ICL, Random ICL, Dynamic ICL, BERT, DistilBert, Roberta, SciBert, Brick-DICL},
    xtick=data,
    x tick label style={rotate=45, anchor=east, font=\scriptsize},
    enlarge x limits=0.08,
]

\nextgroupplot[title={B1 Building}, ybar, /pgf/bar width=5pt, legend style={at={(0.7,0.98)}, anchor=north, font=\small}]
\addplot[fill=blue!40] coordinates {(Prompt,51.49) (Static ICL,53.23) (Random ICL,52.14) (Dynamic ICL,78.31) (BERT,15.78) (DistilBert,5.26) (Roberta,15.79) (SciBert,15.79) (Brick-DICL,90.30)};
\addlegendentry{Stage2 Hits@1}
\addplot[fill=red!40] coordinates {(Prompt,65.43) (Static ICL,64.24) (Random ICL,63.35) (Dynamic ICL,82.32) (BERT,21.05) (DistilBert,10.52) (Roberta,21.05) (SciBert,21.05) (Brick-DICL,96.73)};
\addlegendentry{Stage2 Hits@3}

\nextgroupplot[title={B2 Building}, ybar, /pgf/bar width=5pt, legend style={at={(0.7,0.98)}, anchor=north, font=\small}]
\addplot[fill=blue!40] coordinates {(Prompt,63.15) (Static ICL,67.36) (Random ICL,63.15) (Dynamic ICL,82.63) (BERT,46.22) (DistilBert,30.25) (Roberta,52.10) (SciBert,45.37) (Brick-DICL,96.38)};
\addlegendentry{Stage2 Hits@1}
\addplot[fill=red!40] coordinates {(Prompt,74.89) (Static ICL,77.73) (Random ICL,77.78) (Dynamic ICL,87.89) (BERT,58.82) (DistilBert,47.05) (Roberta,60.50) (SciBert,60.50) (Brick-DICL,98.94)};
\addlegendentry{Stage2 Hits@3}

\end{groupplot}
\end{tikzpicture}
\vspace{-2mm}
\caption{Stage2 Hits@1 and Hits@3 accuracy comparison for BMS point classification on B1 and B2 buildings.}
\label{fig:baseline_comparison}
\end{figure*}

\begin{figure*}[t]
\centering
\begin{tikzpicture}
\begin{groupplot}[
    group style={
        group size=4 by 1,
        horizontal sep=1.0cm,
    },
    width=0.28\textwidth,
    height=4.5cm,
    grid=major,
    grid style={dashed, gray!30},
    label style={font=\small},
    tick label style={font=\scriptsize},
    title style={font=\small},
]

\nextgroupplot[title={B1 (Metadata Shots)}, xlabel={\# of Shots}, ylabel={Accuracy (\%)}, ymin=55, ymax=100, xtick={1,7,13,15}, legend to name=combinedlegend, legend columns=3, legend style={font=\small}]
\addplot[mark=square*, blue, thick] coordinates {(1,61.95) (4,70.65) (7,75.00) (10,72.82) (13,76.08) (15,76.08)};
\addlegendentry{Stage1 Hits@1}
\addplot[mark=triangle*, red, thick] coordinates {(1,83.69) (4,85.86) (7,88.04) (10,88.04) (13,91.30) (15,91.30)};
\addlegendentry{Stage2 Hits@1}
\addplot[mark=o, orange, thick] coordinates {(1,93.46) (4,94.47) (7,95.65) (10,95.65) (13,93.47) (15,95.65)};
\addlegendentry{Stage2 Hits@3}

\nextgroupplot[title={B2 (Metadata Shots)}, xlabel={\# of Shots}, ymin=55, ymax=100, xtick={1,7,13,15}]
\addplot[mark=square*, blue, thick] coordinates {(1,62.85) (4,71.37) (7,76.45) (10,82.45) (13,85.75) (15,90.52)};
\addplot[mark=triangle*, red, thick] coordinates {(1,81.54) (4,85.51) (7,89.76) (10,93.76) (13,93.76) (15,96.38)};
\addplot[mark=o, orange, thick] coordinates {(1,93.54) (4,95.54) (7,96.45) (10,96.45) (13,98.94) (15,98.94)};

\nextgroupplot[title={B1 (Retrieved Classes)}, xlabel={\# of Classes}, ylabel={Accuracy (\%)}, ymin=88, ymax=100, xtick={5,15,20,30}]
\addplot[mark=triangle*, red, thick] coordinates {(5,89.13) (10,89.13) (15,89.13) (20,90.30) (25,89.13) (30,89.13)};
\addplot[mark=o, orange, thick] coordinates {(5,94.56) (10,95.65) (15,95.65) (20,96.73) (25,95.65) (30,95.65)};

\nextgroupplot[title={B2 (Retrieved Classes)}, xlabel={\# of Classes}, ymin=88, ymax=100, xtick={5,15,20,30}]
\addplot[mark=triangle*, red, thick] coordinates {(5,93.54) (10,95.54) (15,96.38) (20,96.38) (25,95.54) (30,95.54)};
\addplot[mark=o, orange, thick] coordinates {(5,95.54) (10,96.45) (15,98.94) (20,98.94) (25,96.45) (30,96.45)};

\end{groupplot}
\node at ($(group c2r1.south)!0.5!(group c3r1.south)+(0,-1.2cm)$) {\ref{combinedlegend}};
\end{tikzpicture}
\vspace{-2mm}
\caption{Impact of hyperparameters on Brick-DICL. Left two: number of shots in metadata ICL. Right two: number of retrieved classes in class ICL.}
\label{fig:hyperparams}
\end{figure*}
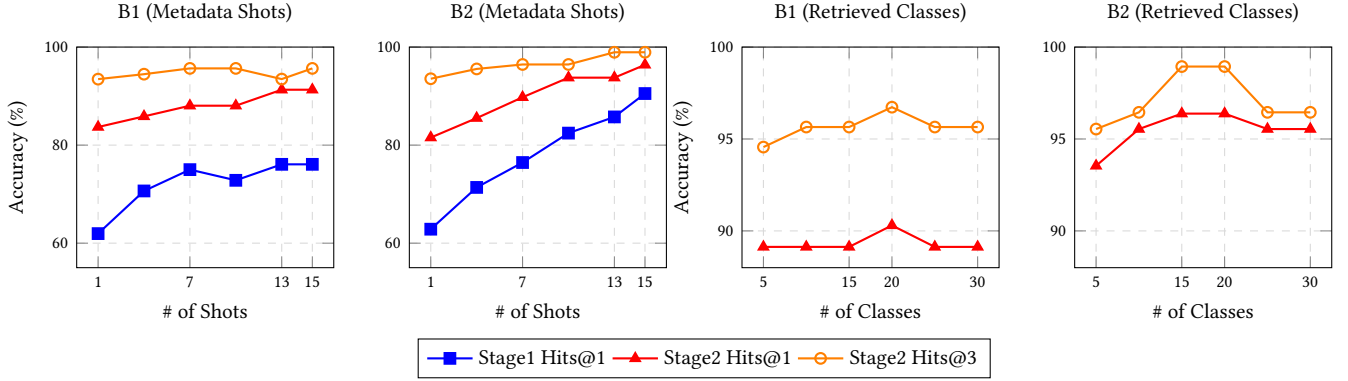

\section{Experiment} In this section, we first introduce two building datasets B1 and B2. Then we conduct extensive experiments to evaluate Brick-DICL on Brick point classification tasks. 

\subsection{Experiment Setup}
\textbf{Datasets.} We evaluate on two BMS datasets from two buildings (B1 and B2). Both datasets contain BMS metadata and ground-truth Brick class labels. The latest Brick schema has 936 point classes and 3,897 parent-class relation pairs. In our approach, each building data is split into training set and testing set. At stage one, we utilize training set to build the metadata knowledge base. At stage two, we utilize all Brick classes to build the class knowledge base.

\textbf{Baseline.} To comprehensively evaluate our Brick-DICL framework, we compare Brick-DICL with eight baseline models in two groups:
\noindent\textbf{A1: prompting-based methods.} We tried four sets of prompting-based methods including standard prompting without few-shot ICL~\cite{sahoo2024systematic}, static ICL~\cite{gao2020making}, random ICL~\cite{rubin2022learning}, and dynamical ICL~\cite{ye2021crossfit}. For static/random/dynamic ICL, we select few-shot input samples from the training set in a static/random/dynamic manner and further ask LLMs to map input samples to the Brick classes.
\noindent\textbf{A2: pre-trained language models.} We fine-tune four pre-trained language models over the building data, including  BERT~\cite{devlin2019bert}, DistilBert~\cite{sanh2019distilbert}, Roberta~\cite{liu2019roberta}, and SciBert~\cite{beltagy2019scibert}.

\textbf{Experimental Setting.} To evaluate Brick-DICL, we adopt the widely-used metrics at each stage: Stage1 hits@1, Stage2 hits@1, and Stage2 hits@3. Hits@1 refers to the accuracy of the model's top generation result being correct, while hits@3 measures the accuracy of the correct classification being among the model's top three generation results. Stage1 hits@1 is the prediction accuracy at stage one, while Stage2 hits@1 and Stage2 hits@3 refer to the classification accuracy at stage two. Mention that, if the predicted class is the parent of the ground-truth class, following the existing work~\cite{balaji2016brick}, we consider it as a correct classification. Besides, we used multiple LLMs to filter low-confidence generations.~\label{sec: exper}

\subsection{Performance Discussion on Brick-DICL}
\subsubsection{Performance Comparison among Baselines}

Figure~\ref{fig:baseline_comparison} summarizes the Hits@1 and Hits@3 accuracy for the BMS point classification task on two buildings, B1 and B2. The results clearly highlight several important trends. Firstly, ICL-based approaches (A1) exhibit a substantial advantage over conventional transformer models (A2). In particular, dynamic ICL achieves notably higher Hits@3 accuracy on both buildings compared to BERT-based models. This demonstrates the strong capability of ICL strategies to leverage context for enhanced semantic understanding.

Secondly, a closer examination within group A1 reveals that dynamic ICL delivers markedly better performance than its static counterpart. This underscores the crucial role of dynamic context adaptation in extracting more relevant, task-specific representations from the data, as opposed to utilizing a fixed context.

Thirdly, among standard transformer models in group A2, RoBERTa achieves the highest accuracy on B2, while DistilBERT shows the lowest performance, suggesting that model architecture and pretraining methodology significantly influence outcomes in this classification scenario.

Finally, Brick-DICL consistently outperforms all baseline methods across both buildings and both metrics, validating the effectiveness and robustness of our proposed dynamic ICL framework for Brick schema-based point classification.

\subsubsection{Impact of Number of Shots in Metadata ICL}

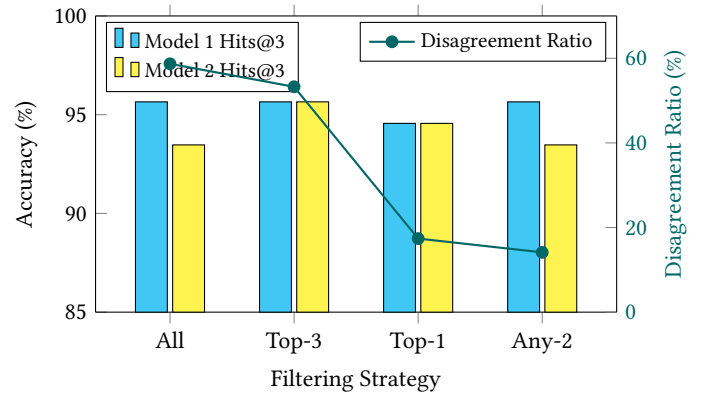
\begin{figure}[b]
\centering
\begin{tikzpicture}
\begin{axis}[
    width=\columnwidth,
    height=5.5cm,
    ybar,
    /pgf/bar width=12pt,
    ylabel={Accuracy (\%)},
    ymin=85, ymax=100,
    symbolic x coords={All, Top-3, Top-1, Any-2},
    xtick=data,
    xlabel={Filtering Strategy},
    legend style={at={(0.02,0.98)}, anchor=north west, font=\small},
    axis y line*=left,
    enlarge x limits=0.2,
]
\addplot[fill=cyan!60] coordinates {(All,95.65) (Top-3,95.65) (Top-1,94.56) (Any-2,95.65)};
\addlegendentry{Model 1 Hits@3}
\addplot[fill=yellow!80] coordinates {(All,93.47) (Top-3,95.65) (Top-1,94.56) (Any-2,93.47)};
\addlegendentry{Model 2 Hits@3}
\end{axis}
\begin{axis}[
    width=\columnwidth,
    height=5.5cm,
    ylabel={Disagreement Ratio (\%)},
    ymin=0, ymax=70,
    symbolic x coords={All, Top-3, Top-1, Any-2},
    xtick=\empty,
    axis y line*=right,
    axis x line=none,
    enlarge x limits=0.2,
    legend style={at={(0.98,0.98)}, anchor=north east, font=\small},
    ylabel style={teal!80!black},
    y tick label style={teal!80!black},
]
\addplot[mark=*, teal!80!black, thick] coordinates {(All,58.69) (Top-3,53.26) (Top-1,17.39) (Any-2,14.13)};
\addlegendentry{Disagreement Ratio}
\end{axis}
\end{tikzpicture}
\vspace{-2mm}
\caption{Comparison of four filtering strategies among Model 1 and Model 2 on B1 building.}
\label{fig:filtering}
\end{figure}

Figure~\ref{fig:hyperparams} (left two panels) shows the accuracy of Brick-DICL with different numbers of shots in metadata ICL on both buildings. From this figure, we can conclude that: (i) Performance generally improves as the number of shots increases, with 15-shot ICL achieving the best overall results across most metrics. (ii) The improvement is more pronounced in B2 compared to B1, indicating that more examples provide greater benefit when the metadata is more diverse. (iii) Mid-range configurations (7-shot and 10-shot) offer competitive Stage2 Hits@3 performance while requiring fewer examples. (iv) B2 consistently outperforms B1 across all shot configurations, indicating building-specific characteristics may influence classification performance. To conclude, we select $k=15$ for stage one. 

\subsubsection{Impact of Number of Retrieved Classes in Class ICL}

Figure~\ref{fig:hyperparams} (right two panels) shows the accuracy of Brick-DICL with different numbers of retrieved classes in class ICL. From this figure, we can conclude that: (i) Class retrieval affects only Stage2 performance as expected, since Stage1 prediction occurs before class retrieval. (ii) For B1, Top 20 retrieved classes achieves the best performance, while for B2, both Top 15 and Top 20 yield optimal results. (iii) Performance improvement plateaus or slightly declines beyond 20 retrieved classes, suggesting that including too many candidate classes may introduce noise in the classification process. (iv) The improvement from Top 5 to Top 20 is more significant in Stage2 Hits@3 than in Stage2 Hits@1, indicating that retrieving more classes primarily helps capture the correct class within the top three predictions. In this work, we select $m=20$ in consideration of the performance and efficiency.

\subsubsection{Comparison of Filtering Strategies in Multi-LLM Consensus}

Figure~\ref{fig:filtering} displays a grouped bar and line chart comparing four filtering strategies---All, Top-3, Top-1, and Any-2---across two large language models (Model 1 and Model 2) on the B1 building dataset. The bars represent Stage2 Hits@3 accuracy for Model 1 and Model 2, respectively; the teal line shows the disagreement ratio for each strategy. Model 1 maintains stable accuracy across all filtering mechanisms, indicating robust and consistent performance, whereas Model 2 exhibits greater fluctuation, reflecting less stability. The disagreement ratio markedly drops from the All and Top-3 strategies to Top-1 and Any-2, demonstrating that Top-1 and Any-2 strategies substantially reduce the number of flagged low-confidence predictions. The Any-2 strategy achieves the lowest disagreement ratio, suggesting it is the most efficient for minimizing manual review while preserving strong classification accuracy.

\subsubsection{Ablation Study}

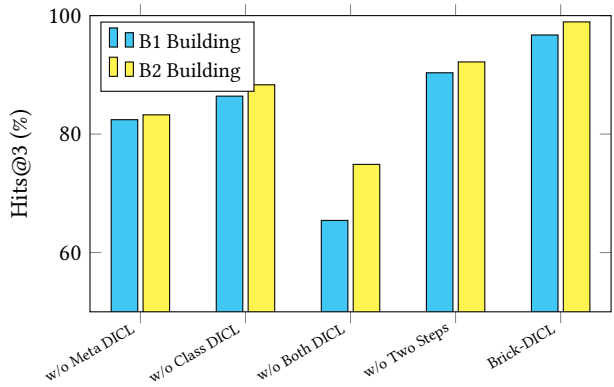
\begin{figure}[t]
\centering
\begin{tikzpicture}
\begin{axis}[
    width=\columnwidth,
    height=5.5cm,
    ybar,
    /pgf/bar width=10pt,
    ylabel={Hits@3 (\%)},
    ymin=50, ymax=100,
    symbolic x coords={w/o Meta DICL, w/o Class DICL, w/o Both DICL, w/o Two Steps, Brick-DICL},
    xtick=data,
    x tick label style={rotate=30, anchor=east, font=\scriptsize},
    legend style={at={(0.02,0.98)}, anchor=north west, font=\small},
    enlarge x limits=0.12,
]
\addplot[fill=cyan!60] coordinates {(w/o Meta DICL,82.43) (w/o Class DICL,86.41) (w/o Both DICL,65.43) (w/o Two Steps,90.35) (Brick-DICL,96.73)};
\addlegendentry{B1 Building}
\addplot[fill=yellow!80] coordinates {(w/o Meta DICL,83.25) (w/o Class DICL,88.32) (w/o Both DICL,74.89) (w/o Two Steps,92.18) (Brick-DICL,98.94)};
\addlegendentry{B2 Building}
\end{axis}
\end{tikzpicture}
\vspace{-2mm}
\caption{Performance comparison among model variants on B1 and B2 buildings.}
\label{fig:ablation}
\end{figure}

Figure~\ref{fig:ablation} presents the results of ablation experiments on Stage2 Hits@3 accuracy for B1 and B2 buildings under five different settings: removing Metadata DICL, Class DICL, Both DICL, Two steps in Stage Two, and using the full Brick-DICL model. The grouped bar chart shows that the removal of either metadata or class DICL individually leads to a moderate decrease in performance, while removing both simultaneously results in the most significant drop in Hits@3 accuracy for both buildings. Excluding the two-step process in Stage Two also reduces accuracy, though not as drastically. The Brick-DICL setting achieves the highest Hits@3 on both B1 and B2 buildings, underscoring the effectiveness of the complete approach. These results demonstrate that each component contributes to overall performance, with the combination of all modules in Brick-DICL being essential for achieving optimal classification accuracy.

\section{Conclusion}
In this paper, we propose Brick-DICL, a framework for automated Brick point classification in building management systems. It consists of two main modules: a two-stage dynamic in-context learning pipeline and a multi-LLM filtering mechanism. The two-stage dynamic ICL learns to map BMS points to standardized Brick schema classes by leveraging metadata RAG followed by class-specific refinement. These learned mappings can be directly applied to building commissioning, energy management, and interoperability tasks. The multi-LLM filtering mechanism automatically identifies low-confidence predictions requiring human review, significantly reducing manual effort while maintaining high accuracy. Comprehensive experiments on multiple building datasets demonstrate Brick-DICL's superiority, achieving high classification accuracy with minimal human intervention. Evaluation of filtering strategies reveals the Any-2 consensus approach achieves the optimal balance between accuracy and disagreement ratio. These results highlight Brick-DICL's capabilities in addressing the challenging task of mapping diverse BMS points to standardized ontologies, accelerating digital building creation.

\bibliographystyle{plain}
\bibliography{amlc_2024}

\end{document}